\documentclass{article}
\usepackage[final]{corl_2022} 

\usepackage{cite}
\usepackage{amsmath,amssymb,amsfonts}
\usepackage{graphicx}
\usepackage{textcomp}
\usepackage{gensymb}
\usepackage{wrapfig}
\usepackage{siunitx}
\usepackage[normalem]{ulem}

\usepackage{comment}
\usepackage{algorithm}
\usepackage{algpseudocode}

\usepackage[dvipsnames]{xcolor}
\definecolor{gray_table}{HTML}{f0f0f5}
\usepackage{colortbl}

\usepackage{pifont}

\newcommand*\colourcheck[1]{%
  \expandafter\newcommand\csname #1check\endcsname{\textcolor{#1}{\ding{52}}}%
}
\colourcheck{ForestGreen}

\usepackage{caption}
\usepackage{subcaption}
\usepackage{enumitem}

\usepackage{tikz}
\usetikzlibrary{external, positioning}
\usepackage{pgfplots}

\usepgfplotslibrary{statistics}
\usepgfplotslibrary{fillbetween}
\usetikzlibrary{shapes}
\usetikzlibrary{er}

\usepackage{titlesec}
\titlespacing{\section}{0pt}{*0.3}{*0.3}
\titlespacing{\subsection}{0pt}{*0.3}{*0.3}
\setlist[itemize]{noitemsep, ,nolistsep,topsep=0pt}
\setlist[enumerate]{noitemsep,nolistsep, topsep=0pt}

\pgfplotstableread[col sep=comma]{images/traj_new.csv}\data

\begin{document}

\title{In-Hand Gravitational Pivoting Using Tactile Sensing}

\author{
  Jason Toskov\\
  Monash University\\
  \texttt{jtos0003@student.monash.edu} \\
  \And
  Rhys Newbury\\
  Monash University\\
  \texttt{rhys.newbury@monash.edu} \\
  \And
  Mustafa Mukadam\\
  Meta AI\\
  \texttt{mukadam@fb.com}
  \And
  Dana Kuli\'{c}\\
  Monash University\\
  \texttt{dana.kulic@monash.edu} \\
  \And
  Akansel Cosgun\\
  Monash University\\
  \texttt{akansel.cosgun@monash.edu} \\
}
\maketitle

\begin{abstract}
We study gravitational pivoting, a constrained version of in-hand manipulation, where we aim to control the rotation of an object around the grip point of a parallel gripper. To achieve this, instead of controlling the gripper to avoid slip, we \emph{embrace slip} to allow the object to rotate in-hand. We collect two real-world datasets, a static tracking dataset and a controller-in-the-loop dataset, both annotated with object angle and angular velocity labels. Both datasets contain force-based tactile information on ten different household objects. We train an LSTM model to predict the angular position and velocity of the held object from purely tactile data. We integrate this model with a controller that opens and closes the gripper allowing the object to rotate to desired relative angles. We conduct real-world experiments where the robot is tasked to achieve a relative target angle. We show that our approach outperforms a sliding-window based MLP in a zero-shot generalization setting with unseen objects. Furthermore, we show a 16.6\% improvement in performance when the LSTM model is fine-tuned on a small set of data collected with both the LSTM model and the controller in-the-loop. Code and videos are available at \url{https://rhys-newbury.github.io/projects/pivoting/}
\end{abstract}


\begin{figure}[htb]
    \centering
    \includegraphics[width=0.5\linewidth]{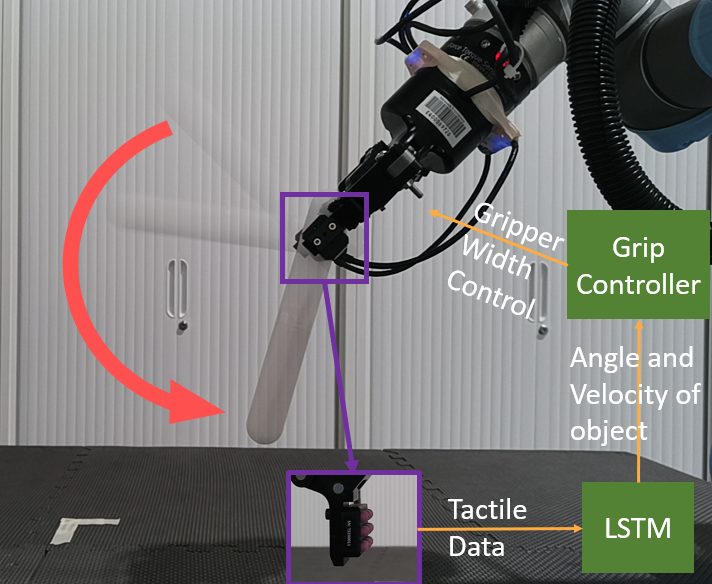}
    \caption{We study in-hand manipulation to rotate an object with gravitational pivoting. We design an LSTM model to predict the position and velocity of an object purely from tactile sensing. Our grip controller uses this prediction to modulate the width of the gripper and achieve a target angle.}
    \label{fig:intro}
\end{figure}

\section{Introduction}

The majority of past works in robotic manipulation either assume fixed grasps~\citep{grasping_survey, grasping_survey2} or aim to avoid any slipping during manipulation~\citep{TactileSurvey}. However, we focus on \emph{embracing slip}, using object slip to increase the dexterity of simple grippers. This idea was explored by \citet{Chen2021_Velocity} where the grip on the object is loosened to allow an object to slip downwards with gravity. This paper explores a method of inducing rotation for in-hand manipulation using gravity. Such manipulation, known as pivoting, is key to performing tasks requiring an object to be at a specific relative angle to a gripper, such as stacking shelves~\citep{9391990}.

This paper proposes a method for a robot with a parallel gripper to rotate a long object grasped away from its center of mass to a desired final relative orientation. This aims to allow parallel grippers to robustly reorient objects into a desired orientation without having to regrasp the object. To achieve this task, we focus on addressing two challenges: tracking the position of the object and controlling the gripper to allow for gravitational pivoting towards the target angle.  

Vision-based methods to track the object often make use of an eye-in-hand camera. However, the gripper will often occlude the object, making it difficult to estimate the angle of the object accurately~\citep{9341799}. An alternative is to use an externally placed camera. However, this necessitates the robot moving to a fixed position in front of the camera for each manipulation. We use purely tactile information to track the object to avoid these issues. We design an LSTM-based neural network model, RSE-LSTM, which uses tactile information to predict a held objects' relative angular position and angular velocity.

Previous approaches to controlling the gripper often used model-based approaches, which required information about the object, such as shape, mass, and friction. In contrast, we design a simple gripper controller that assumes no a priori knowledge about the object parameters to allow for generalization to unseen objects.

We collect a real-world force-based tactile dataset, on ten household objects. This dataset is annotated with both angular position and velocity measurements. RSE-LSTM is trained on this dataset, and the results are reported with respect to both unseen data and unseen objects. We further validate our approach experimentally on unseen objects.

The contributions of our paper are threefold:
\begin{itemize}[leftmargin=*,topsep=0pt,itemsep=0pt]
    \item An annotated dataset containing gravitational pivoting with 10 household objects.
    \item A LSTM-based neural network which can predict both the velocity and angle of an object using only tactile information. 
    \item A grip controller, which can adjust the width of the gripper to allow an object to pivot in-hand to achieve a required relative angle. 
\end{itemize}

\section{Related Works}
\label{sec:related_works}

\textbf{Slip measurement.} Slip detection is often framed as a binary problem, with machine learning models predicting either slip or no slip. This is achieved with the use of visual sensors~\citep{dong2018maintaining}, force-based sensors~\citep{9196615}. or optical sensors~\citep{8403292}. Various machine learning techniques have been used including: Support Vector Machines~\citep{8324455, 8403292}, MLPs~\citep{7363558} and LSTM models~\citep{8461117}. Alternatively, Convolutional Neural Networks (CNN) have been used to both detect and classify the type of slip as either translational or rotational~\citep{Meier2016TactileCN}. LSTM models have been used to determine the direction of rotational slip~\citep{8868431}, or the overall direction of the combination of rotational and translational slip~\citep{zapata2019learning}. 

The domain of quantitative slip measurement is comparatively underexplored. Previous works measure the amount of translational slip using image-based tactile sensors~\citep{doi:10.1177/1729881419846336}. Alternatively, visual gel-based tactile sensors have been used to measure the rotation angle using a model-based approach~\citep{kolamuri2021improving}. However, to our knowledge, the use of force-based tactile sensing has not been explored, which is the focus of this paper. 

\textbf{Induced rotation.} To induce rotation in a held object, previous work has made use of the external environment to apply a torque or force on the held object~\citep{7354264,chavandafle2017samplingbased,ReorientPivot}. However, rotation can also be induced without any interaction with external objects. For example, the robot can perform a swinging motion using the end-effector, where the velocity of the swing aims to bring an object to a desired angle~\citep{8202299, rl_pivoting, Sintov2016}. 

Alternatively, by loosening the grip on the object, gravity can be used to induce a rotation in the object~\citep{7354177, 7487159, ModelBasedPivot, ModelBasedSliding, CAVALLO2020108875_slider, Holladay2015}. These approaches are model-based and rely on prior knowledge of important parameters of the system, such as shape, mass and friction of the held object. Our work assumes no prior knowledge about such parameters to allow for generalization to unseen objects.

\textbf{Induced translational slip.} \citet{Shi2017_Sliding} design a model-based approach to induce translational slip in an object by accelerating the gripper causing the object to slide in a desired way in-hand. \citet{Chen2021_Velocity} train a MLP to predict the velocity of an object which is undergoing translational slip. They feed the MLP the previous one second of observations to predict the sliding velocity of the object. A controller is then designed to achieve a target sliding velocity for the object. Our work extends this to the rotational case and makes use of an LSTM rather than providing a fixed length history. The allows the network to learn an encoding for the history, which may be more informative.

\section{Problem Definition}
\label{sec:problem}
Gravitational pivoting is a form of in-hand manipulation, where an object is rotated in hand by loosening the grip on the object. Consider a static gripper which holds an object away from its center-of-mass. If the gripper is closed tightly enough, the object should remain stationary inside the gripper. However, if the grip on the object is loosened slightly, gravity will induce a torque on the object, causing it to pivot inside the gripper. 


\begin{figure}[t]
    \centering
    \includegraphics[width=0.65\linewidth]{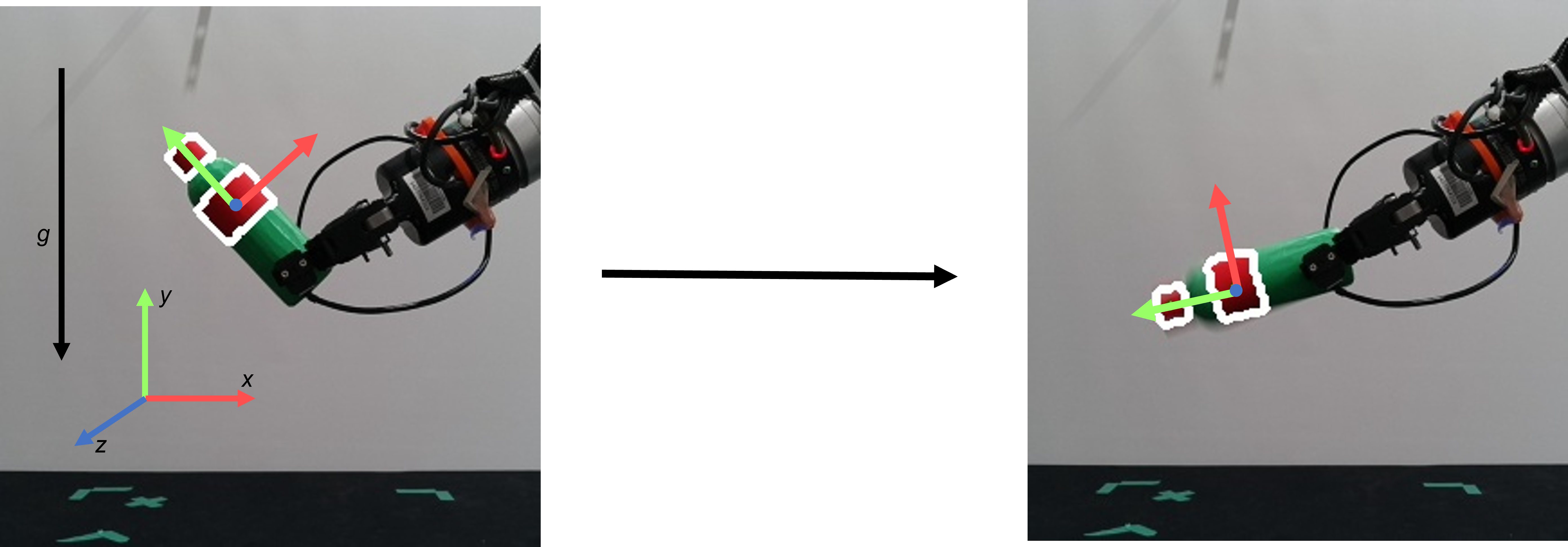}
    \caption{Gravitational Pivoting occurs when a parallel-jaw gripper loosens its grip on a object such that gravity induces a torque on the object. This torque will then cause a rotation to occur. }
    \label{fig:probelm_setup}
\end{figure}

The coordinate system for this problem is defined in Figure \ref{fig:probelm_setup}. Gravity is defined to work in the negative $y$-direction in the global frame. We also define a rotating object-centric coordinate frame, where the $y$-axis is aligned with the long axis of the object. 

\begin{wrapfigure}[17]{R}{0.42\textwidth}
    \includegraphics[width=0.38\textwidth]{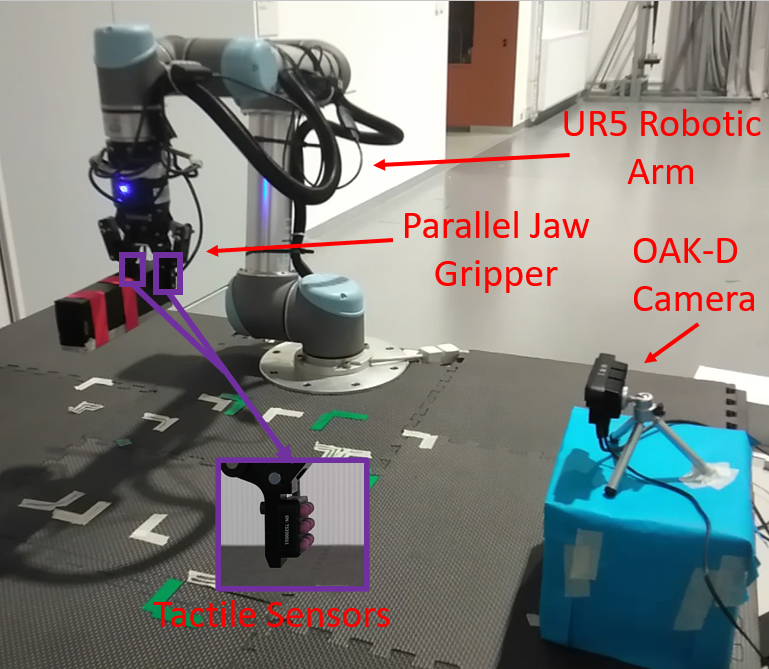}
    \caption{The hardware setup for both data collection and experiments.}
    \label{fig:hardware}
\end{wrapfigure}

Specifically, we consider pivoting tasks as follows. The robot starts with an object in hand, grasped away from the center-of-mass. The task is to rotate the object by a relative angle around the object-centric $z$-axis. We only consider achieving relative angles ($\alpha$) in the range $[0,180]$ degrees, as we only consider allowing the object to fall and rotate. We assume that the robot has no a priori knowledge about the object and \emph{only} has access to tactile information from the fingertips. This aims to allow generalizability by assuming no knowledge of the grasped objects. We constrain the type of objects to ones that have a prism-like shape, where one dimension, length, is much larger than the other two, width and depth. When these objects are held away from the center-of-mass, the torque induced on the object by gravity will be much larger than the downwards force at the contact points. Therefore, the objects are more likely to undergo rotational slip when the grip is loosened, and the translational slip is assumed to be negligible. Prism-like objects include common household objects, such as bottles, boxes, and tools (such as hammers).

\section{Data Collection}
\label{sec:datacollection}
\subsection{Hardware Setup}

The robot hand consists of a Robotiq 2f-85 parallel-jaw gripper and a table-mounted UR5 robot arm. On each jaw of the gripper is a PapillArray tactile sensor~\citep{Khamis2018PapillArrayAI}. The parallel jaw gripper has a width of 85mm and can be controlled in 256 increments, for a resolution of 0.33mm per increment. The PapillArray sensors consist of 9 pillars, arranged in a 3 by 3 square, where each pillar provides a force and displacement measurement in each direction, and whether the pillar is in contact with the object. The sensor also provides global forces and torques in each direction, for a total of 142 measurements over the two sensors. In addition, an OAK-D camera is positioned to provide a side-on view of the robotic arm and held object, which will be used to record the ground truth angle (more details of the ground truth collection is provided in Section \ref{sec:datacoll}). The hardware setup for both data collection and experiments is shown in Figure \ref{fig:hardware}.

\subsection{Object set}

\begin{wrapfigure}[18]{r}{0.5\textwidth}
    \vspace{-\intextsep}
    \includegraphics[width=0.48\textwidth]{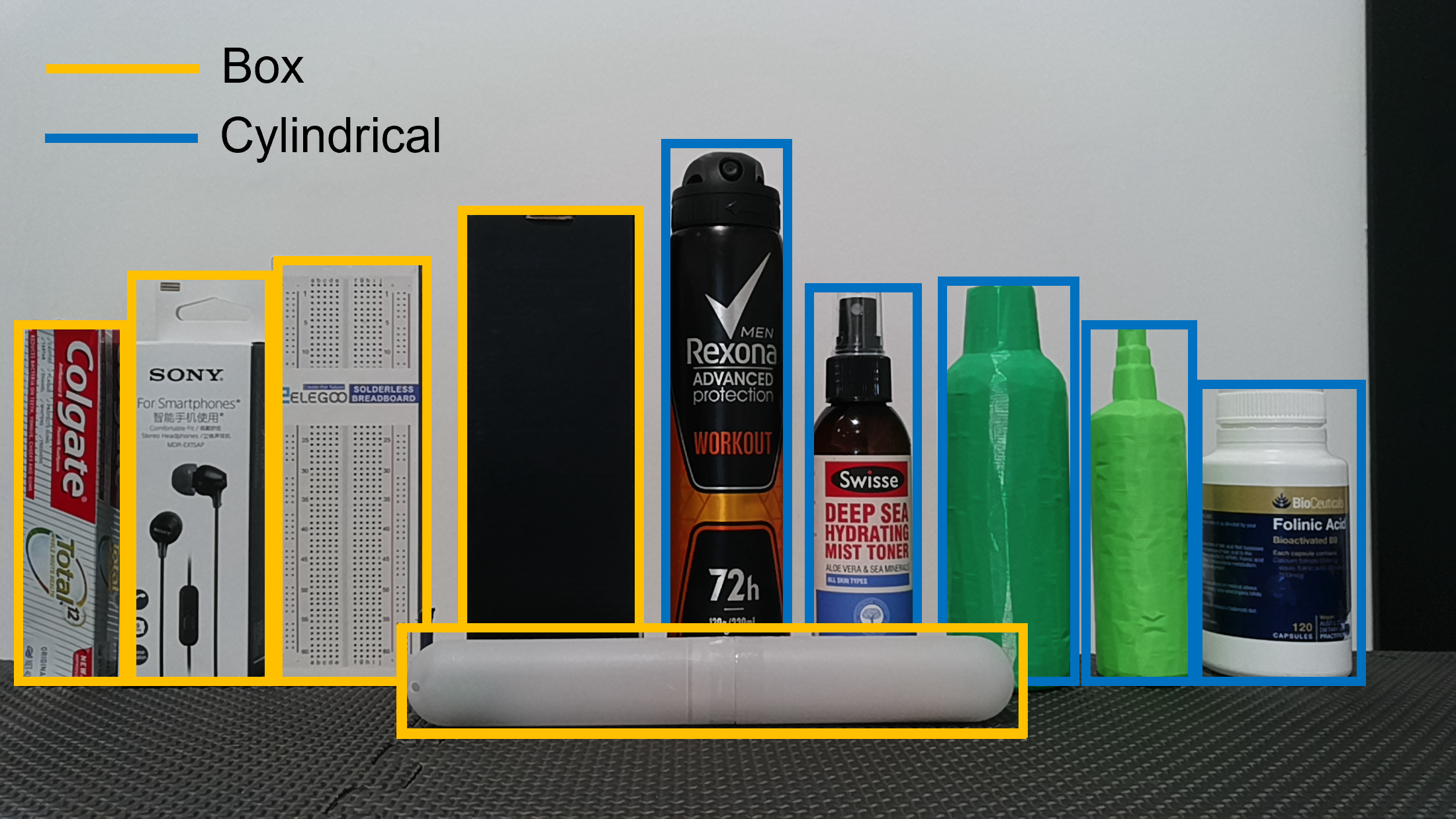}
    \caption{Objects used in this dataset. We distinguish between two classes of objects, box-like object and cylinder-like objects. We refer to the names of these objects throughout paper. The names are from left to right (back row): Toothpaste, Earbud, Breadboard, Magnet, Deodorant, Spray2, Shampoo, Spray1, Pill. The object at the front is labeled as Toothbrush.}
    \label{fig:object_set}
\end{wrapfigure}

We consider the pivoting task for a set of 10 different household objects. Similar to the object set used by~\citet{Chen2021_Velocity}, we consider two classes of objects: box-like objects and cylinder-like objects, with five objects in each class. The set of objects used are shown in Figure \ref{fig:object_set}. 

\subsection{Methodology}
\label{sec:datacoll}

A systematic methodology is used to collect a dataset of gravitational pivoting, outlined in the supplementary material. We use two different methodologies for controlling the gripper during rotation of the object:

\begin{itemize}[leftmargin=*,topsep=0pt,itemsep=0pt]
    \item \textbf{Rotate To Stop}: The gripper is opened a fixed amount and the object is allowed to rotate until the object comes to a stationary position.
    \item \textbf{Angle Goal}: The controller (described in Section \ref{subsec:gripcontroller}) is tasked to stop the object at an angle ($\alpha_{stop}$) from a set of angles ($\mathrm{A}_{stop}$). The controller receives ground truth angle readings for the purposes of data collection.  
\end{itemize}

To create a larger variation of friction properties, we collect data both with and without a layer of masking tape added to each object surface. In total there were 595 `Rotate To Stop' and 971 `Angle Goal' sequences collected, after filtering invalid datapoints. The dataset is attached in the supplementary material.


\subsection{Ground Truth Annotation}

To measure the ground truth rotation of the grasped object, two distinctly colored blobs have been attached to each object. An external camera observes these blobs, and we then define the object-centric $y$-axis between the centroid of the two blobs. The orientation change between the initial object-centric $y$-axis and the current object-centric $y$-axis is used as the ground-truth angles. The position and angular velocity are filtered to ensure they follow a smooth signal. The details of the filter are in the supplementary material.


\section{Proposed Approach}
\label{sec:approach}

Our approach consists of two main components. A Rotational Slip Estimator LSTM (RSE-LSTM) and a Grip Controller. The system diagram is shown in Figure \ref{fig:intro}. From purely tactile information, the RSE-LSTM estimates the relative angle change between the initial and current object-centric $y$-axis. The Grip Controller uses the estimation and gravitational pivoting, aiming to reach a desired rotation relative to the initial angle. 

\subsection{Rotational Slip Estimator}

The RSE-LSTM uses measurements from the tactile sensors and predicts both the current angular velocity ($\omega$) and relative angle of the object ($\alpha$) as the outputs. We found that calculating both $\alpha$ and $\omega$ improved the results of the model (Section \ref{subsec:rotest}).

The RSE-LSTM model consists of an LSTM, the outputs of which are passed to an MLP. Instead of using a sliding window (similar to \citep{Chen2021_Velocity}), using the hidden states of the LSTM could allow the model to use a longer history more effectively by learning a more feature-rich hidden state. The model runs at 60 Hz due to hardware limitations in training data collection.  

\subsection{Grip Controller}
\label{subsec:gripcontroller}

\begin{algorithm}[htb]
\caption{Grip Controller}\label{alg:cap}
\begin{algorithmic}
\State $goal \gets \alpha_{stop}$
\State $error \gets relative\_goal$
\While {$error > \epsilon_{\alpha}$}
\State $m \gets \textrm{Observation()}$
\State $\alpha, \omega \gets \textrm{RSE-LSTM}(m)$
\If {$\omega < \omega_{min}$ and $t_{curr} - t_{prev} > t_{wait}$}
\State $\textrm{increase\_gripper\_opening()}$
\State $t_{prev} \gets t_{curr}$
\EndIf
\State $\alpha_F \gets \textrm{fp}(\alpha, \omega, d)$
\State $error \gets |\alpha_F - goal|$
\EndWhile
\State {fully\_close\_gripper()}
\end{algorithmic}
\end{algorithm}

The grip controller algorithm is described in Alg.\ref{alg:cap}. The grip controller uses the position and velocity estimated by the LSTM to achieve a target angle. While the object is not moving, the controller slightly opens the gripper. Once the object reaches the target angle, the gripper then closes. However, this alone constantly overshoots the target due to the delay in closing the gripper. To account for this, the velocity estimate is used to forward predict the angle (denoted by fp in Alg. \ref{alg:cap}), by a fixed delay ($d$) to stop the gripper early. The forward prediction equation is: $\alpha_F = \alpha + d \times \omega$.


\section{Pose Estimate Model Evaluation}
\label{sec:trainingresuts}
\subsection{Baseline}
\label{subsec:trainbaseline}
\citet{Chen2021_Velocity} found promising results using a sliding-window MLP in a similar task, where they predict the velocity of an object undergoing translational slip. Therefore, as a baseline, we consider the use of a sliding window \textbf{MLP}, which takes the last 15 steps of tactile data and predicts the angle and angular velocity at the next timestep. The parameters of this model were optimized using a sweep to maximize the performance on the test set, during random split experiments. In Section \ref{subsec:windowexperiments} we present results an ablation study on the length of the sliding window.

\subsection{Experiments}
\label{subsec:trainexperiments}

\begin{wrapfigure}[14]{r}{0.6\textwidth}
    \centering        
    \resizebox{.65\textwidth}{!}{
    \begin{tikzpicture}
    
    \begin{axis}[
        name=mainplot,
        xmin = 0.0,
        xmax= 3,
        ymin = -200.0,
        ymax = 900,
        enlarge y limits=false,
        width = 0.9\textwidth,
        height = 0.45\textwidth,                
        xlabel= time (s),
        ytick={-100, 0, 250, 500, 750},
        legend style={nodes={scale=0.75, transform shape}},
        legend pos  = north east,
        ]
        
        \addplot[red, mark=none] table[x=t , y=velocity] {\data};
        \addplot[blue, mark=none] table[x=t , y=position] {\data};
        

        \addplot[mark=none, draw=none, name path=A] coordinates {(0,-1000) (0.6,-1000)};
        \addplot[mark=none, draw=none,name path=B] coordinates {(0,1650) (0.6,1650)};
        \addplot[red, fill opacity=0.2] fill between[of=A and B,soft clip={domain=0:2.5}];

        \addplot[mark=none, draw=none, name path=A] coordinates {(0.6,-1000) (1.7,-1000)};
        \addplot[mark=none, draw=none,name path=B] coordinates {(0.6,1650) (1.7,1650)};
        \addplot[green, fill opacity=0.2] fill between[of=A and B,soft clip={domain=0:3.5}];
        
        \addplot[mark=none, draw=none, name path=A] coordinates {(1.7,-1000) (3.5,-1000)};
        \addplot[mark=none, draw=none,name path=B] coordinates {(1.7
        ,1650) (3.5,1650)};
        \addplot[blue, fill opacity=0.2] fill between[of=A and B,soft clip={domain=0:3.5}];

        \node[] at (axis cs: 0.3,750) {Initial State};
        \node[] at (axis cs: 1.0,750) {Dynamic Range};
        \node[] at (axis cs: 2,750) {Steady State};


        \legend{
            $\omega (\degree/s)$, 
            $\alpha (\degree)$
        }             
        
    \end{axis}
        
        

 
\end{tikzpicture}
    }
    \caption{We split gravitational pivoting in to three sections, denoted by the three colors in the graph}
    \label{fig:sections}
\end{wrapfigure}
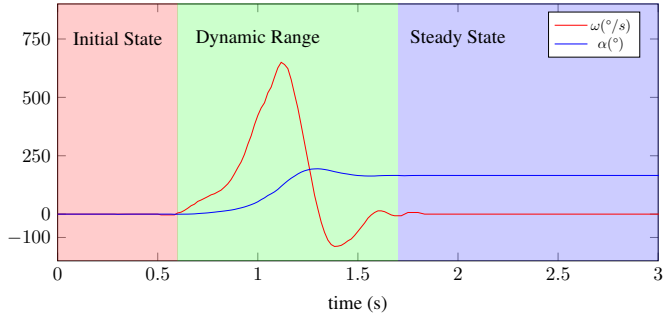

We split the trajectory in to three sections (shown in Figure \ref{fig:sections}).

\begin{enumerate}[leftmargin=*,topsep=0pt,itemsep=0pt]
    \item Initial State (IS): The object is held still, before any movement. The networks should be able to predict both 0 for the initial relative angle and 0 for velocity.
    \item Dynamic Range (DR): The object is currently rotating. The models should track both position and velocity. 
    \item Steady State (SS): The object has finished rotating. The network should be able to predict the correct final angle and 0 velocity.
\end{enumerate}  

The dynamic range commonly lasts less than a second, thus necessitating learned estimation and control work in real-time. The objects also commonly reach velocities of up to 750$\degree$/s, making accurately predicting the velocities of the object difficult. The fast rotation of the object also increases the need for both the perception and control to be real-time, as any time delay in the system will magnify any errors.


We conduct three experiments, firstly experimenting on unseen data (Section \ref{subsec:rotest}), secondly experimenting on unseen objects (Section \ref{subsec:unseenobjtrain}) and finally experimenting on unseen classes. The training and network parameters are detailed in the supplementary material with the unseen classes experiment results.

\begin{figure}[htb]
    \centering
    \pgfplotstableread[col
sep=comma]{images/tracking_examples/pos_1.csv}\posdata

\pgfplotstableread[col
sep=comma]{images/tracking_examples/pos_2.csv}\otherposdata

\pgfplotstableread[col
sep=comma]{images/tracking_examples/vel_1.csv}\veldata

\pgfplotstableread[col
sep=comma]{images/tracking_examples/vel_2.csv}\otherveldata

\begin{tikzpicture}
    \begin{axis}[
        xmin = 0.0,
        xmax= 3.85,
        ymin = 0,
        ymax = 120,
        enlarge y limits=false,
        width = 0.45\textwidth,
        height = 0.4\textwidth,                
        xlabel= time (s),
        ylabel=Angle ($\degree$),
        ytick={30, 60, 90, 120},
        legend style={nodes={scale=0.75, transform shape}},
        legend pos  = north west,
        ]
        
        \addplot[red, mark=none] table[x=t2 , y=gt2] {\otherposdata};
        \addplot[blue, mark=none] table[x=t2 , y=mlp2] {\otherposdata};
        \addplot[green, mark=none] table[x=t2 , y=lstm2] {\otherposdata};
        
        \legend{
            Ground Truth, 
            MLP,
            LSTM,
        }             
        
            \end{axis}

\end{tikzpicture}
    \begin{tikzpicture}
    \begin{axis}[
        xmin = 0.0,
        xmax= 3.85,
        ymin = 0,
        ymax = 550,
        enlarge y limits=false,
        width = 0.45\textwidth,
        height = 0.4\textwidth,                
        xlabel= time (s),
        ylabel=Angular Velocity ($\degree$/sec),
        ytick={100, 200, 300, 400, 500},
        legend style={nodes={scale=0.75, transform shape}},
        legend pos  = north west,
        ]
        
        \addplot[red, mark=none] table[x=Time , y=GT] {\otherveldata};
        \addplot[blue, mark=none] table[x=Time , y=MLP] {\otherveldata};
        \addplot[green, mark=none] table[x=Time , y=LSTM] {\otherveldata};
        
        
            \end{axis}

\end{tikzpicture}
    
    \caption{Example Tracking Performance. The LSTM produces a much smoother output compared to the MLP. The models display a consistent undershooting of the peak angular velocity.}
    \label{fig:examples}
\end{figure}
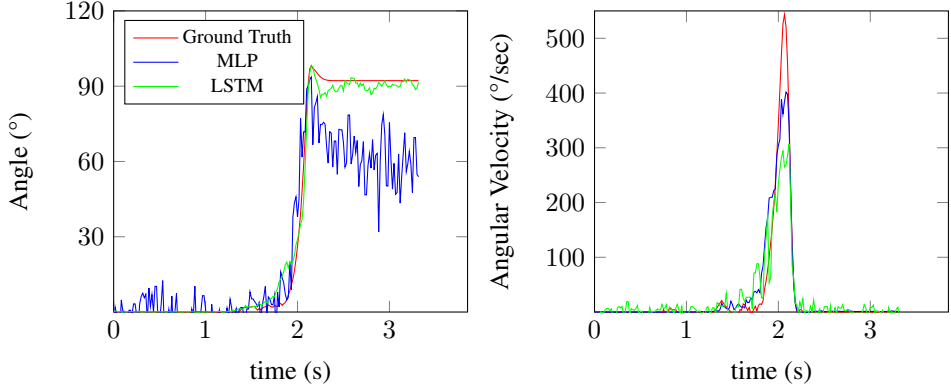

\subsection{Rotation Estimation}
\label{subsec:rotest}

We train the previously described \textbf{LSTM} and \textbf{MLP} on the collected dataset. We varied the output of both models, to predict only $\alpha$, only $\omega$ or both $\omega$ and $\alpha$. If the model only outputs $\omega$, $\alpha$ will be recovered by integrating the signal. If the model calculates only $\alpha$, $\omega$ will be calculated by taking the derivative. An 80/20 train/test split of the data was used for training and the mean absolute error (MAE) is reported in Table \ref{tab:unseen}.

\begin{table}[htb]
\label{tab:unseen}
\begin{tabular}{l|lll|lll}
 & \multicolumn{3}{c}{Angular Error ($\degree$)} & \multicolumn{3}{|c}{Angular Velocity Error ($\degree/s$)}                     \\ \cline{2-7} 

     & \multicolumn{1}{c}{IS}      & \multicolumn{1}{c}{DR}      & \multicolumn{1}{c|}{SS}      & \multicolumn{1}{c}{IS}     & \multicolumn{1}{c}{DR}      & \multicolumn{1}{c}{SS}      \\ \hline
\textbf{LSTM Both} & 0.29±0.06 & \textbf{4.39±0.18}  & \textbf{7.23±0.2}   & \textbf{0.64±0.22} & 32.38±2.6  & 1.47±0.42 \\
\textbf{MLP Both}  & 0.67±0.58 & 10.58±0.57 & 19.85±1.07 & 0.76±0.27 & 41.38±1.64 & 1.37±0.3 \\
\textbf{LSTM ($\omega$-only)}  & \textbf{0.27±0.03} & 7.4±2.1 & 13.79±4.36 & 0.71±0.20 & \textbf{31.97±2.36} & \textbf{0.81±0.26} \\
\textbf{MLP ($\omega$-only)}  & 0.39±0.11 & 12±1.14 & 22.21±2.35 & 1.38±0.5 & 36.93±2.01 & 1.92±0.57 \\
\textbf{LSTM ($\alpha$-only)}  & 0.31±0.03 & 4.52±0.21 & 7.23±0.26 & 1.74±0.32 & 40.13±1.86 & 4.8±0.25 \\
\textbf{MLP ($\alpha$-only)}  & 1.48±0.55 & 10.41±0.67 & 19.79±1.11 & 3.52±1.28 & 65.07±1.62 & 8.31±0.81 \\
\end{tabular}
\caption{Testing results on unseen data. Each network is trained 5 times. IS, DR, SS represent the three sections of motion, Initial State, Dynamic Range and Steady State, respectively. $\omega$ represents the angular velocity, and $\alpha$ represents the angular position.}
\label{tab:unseen}
\end{table}

\textbf{LSTM Both} out-performed the other models in terms of the position error, slightly outperforming \textbf{LSTM ($\alpha$-only)} in terms of the position error in all three stages. \textbf{LSTM Both} also performed similarly to \textbf{LSTM $\omega$-only} in terms of the velocity error. The MLP performance was lower than the LSTM for all stages for both velocity and position error. 


An additional benefit of the LSTM over the MLP is that the LSTM produced much smoother results than the MLP. This can be seen in Figure \ref{fig:examples}, where the MLP predictions are more noisy compared to LSTM. This is problematic when paired with the Grip Controller, making it challenging for the controller to achieve precise angles and also more prone to early failure.

\newpage
\subsection{Unseen Objects}
\label{subsec:unseenobjtrain}

\begin{wrapfigure}[15]{R}{0.55\textwidth}
    \centering
    \includegraphics[width=0.55\textwidth]{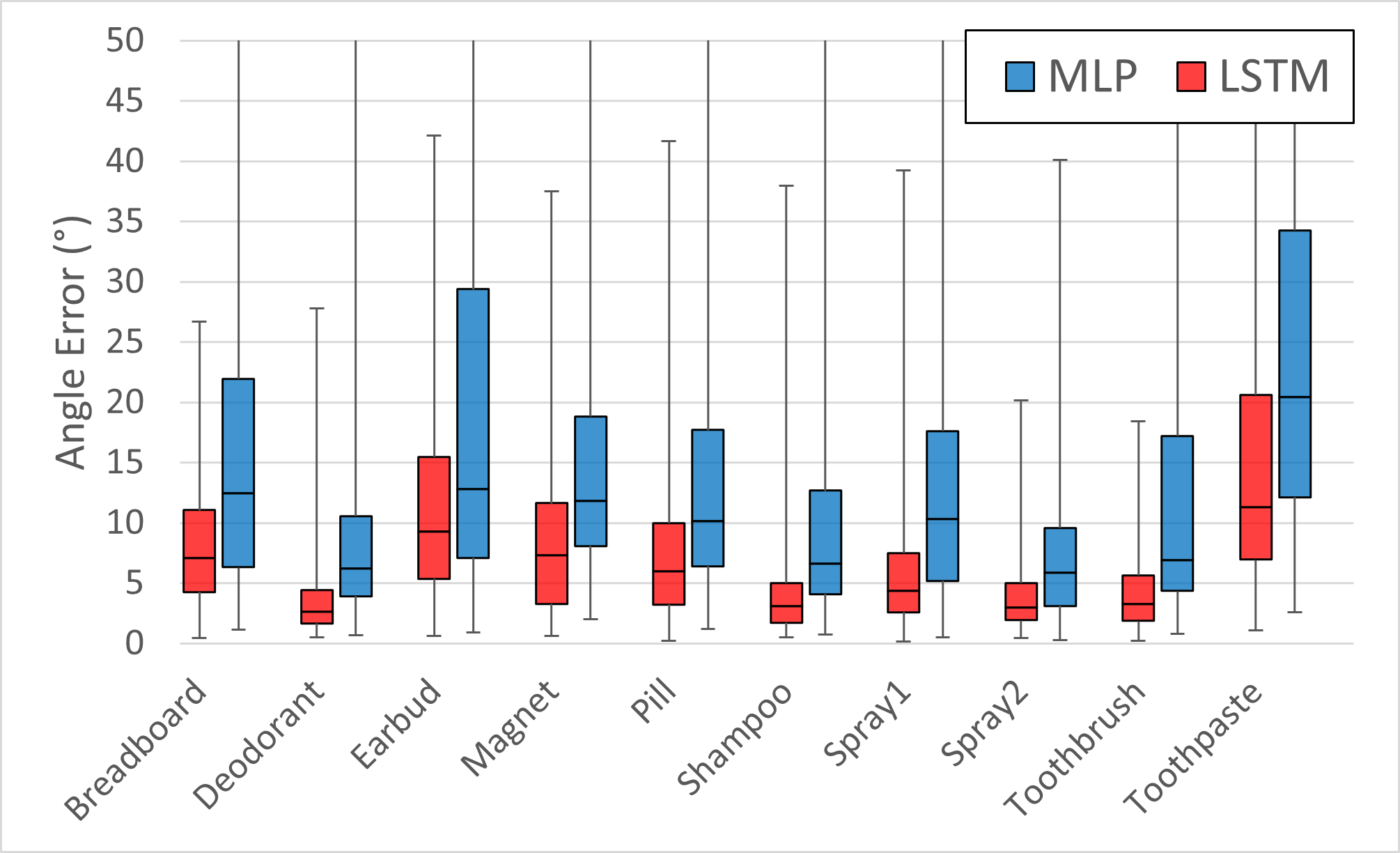}
    \caption{Distribution of position errors for each\\unseen object.}
    \label{fig:histgram}
\end{wrapfigure}

For the next experiments, we only consider models that output both $\alpha$ and $\omega$, as they were the best performing models. We investigate the performance of the models when generalized to unseen objects. We trained the LSTM and MLP on 9 objects and tested on the remaining 1 unseen object. The performance of the models decreased compared to training on previously seen objects. This is most likely due to the model struggling to generalize to objects that have unique properties compared to the other objects in the training set. An example is the `Toothpaste' object, where the position error is much larger for both the LSTM and MLP, as shown in Figure \ref{fig:histgram}.

\section{Controller Evaluation}
\label{sec:experiments}
\subsection{Baselines}

We consider two baselines:
\begin{itemize}[leftmargin=*,topsep=0pt,itemsep=0pt]
    \item \textbf{Oracle}: Ground truth angle and angular velocity measurements for the Grip Controller are taken from visual data. This provides an upper bound on the performance of the gravitational pivoting system.
    \item \textbf{MLP}: The MLP described in Section \ref{subsec:trainbaseline}
\end{itemize}

\subsection{Experimental Setup}

The experimental methodology followed was same as the `Angle Goal' data collection process described in Section \ref{sec:datacollection}.

We experimentally tested the performance on unseen objects using the models from Section \ref{subsec:unseenobjtrain}. During experiments both the ground truth and predicted $\alpha$/$\omega$ were recorded. These measurements were used to compute the MAE of the model within the 3 sections as defined in Section \ref{subsec:trainexperiments} (IS, DR and SS). 

The most important error in the controller experiments is the error in the final angle achieved by the controller, denoted as \textbf{target error (TE)}. We also report the \textbf{failure rate (FR)} which is the percentage of trials for which the controller either dropped the object or became stuck, predicting an $\omega > \omega_{min}$ while the object is stationary.

To address the model generalization to the real-world setup, we further collected a small amount of `Angle Goal' data with the controller receiving angle measurements from the trained RSE-LSTM model instead of the Oracle to test the effects of fine-tuning the controller on the desired scenario. The data collection parameters are detailed in the supplementary material.


\newpage
\subsection{Results}

\begin{wrapfigure}[15]{R}{0.55\textwidth}
    \centering
    \vspace{-\baselineskip}
    \includegraphics[width=0.55\textwidth]{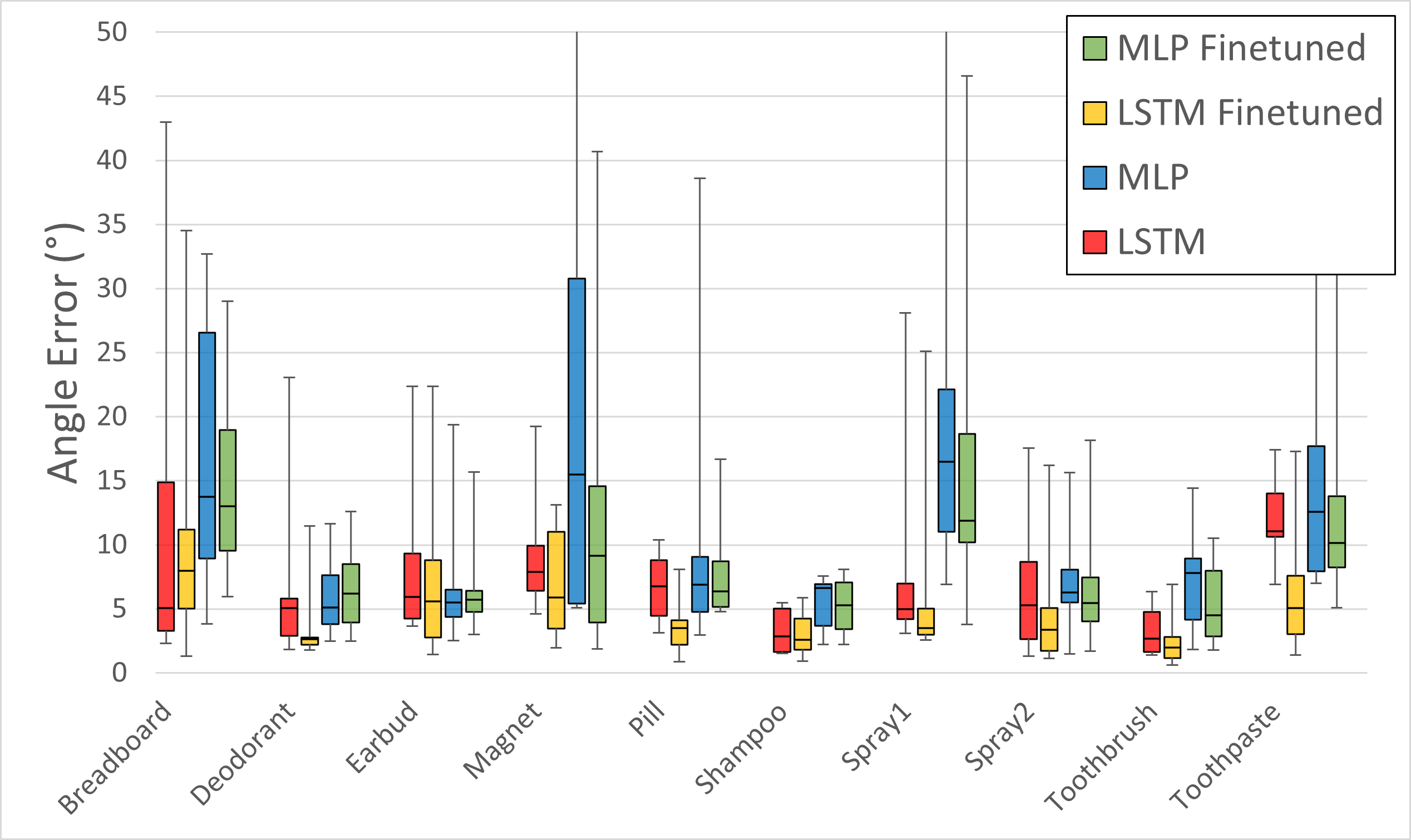}
    \caption{The target-error for all unseen object experiments. The fine-tuned LSTM outperformed all other methods for 8 out of the 10 objects.}
    \label{fig:histgram_finetuned}
\end{wrapfigure}

The results of the real-world experiments are shown in Table \ref{tab:finetune_controller}. While the LSTM outperformed the MLP for all metrics, the tracking error was much larger than the error seen on the test set. This is likely because the training dataset does not represent all scenarios seen in real-world testing such as overshooting the target angle and the gripper not closing.

Furthermore, the target error was much larger than the \textit{Oracle} baselines. This can be attributed to the previously stated tracking inaccuracies, particularly in dynamic range, resulting in the model predicting the target angle was reached either too late or too early.

As shown in Figure \ref{fig:histgram_finetuned}, fine-tuning improved the results on most objects. The distribution of the box plot follows a similar pattern to Figure \ref{fig:histgram}. This suggests some objects are harder for our approach to generalize to. This could be further improved by collecting data on a wider range of objects. 

\begin{table}[tbh]
\centering
\begin{tabular}{l|ll}
 & Failure Rate (\%) & Target Error (\degree) \\ \hline
\rowcolor{gray_table}
\textbf{Oracle} & 0 & 3.95±3.02 \\
\textbf{LSTM} & \textbf{3.3} & 15.2±4.69 \\
\textbf{MLP}  & 6.7 & 23.08±8.31 \\
\rowcolor{gray_table}
\textbf{LSTM-finetuned} & 4.44 & \textbf{12.67±5.31} \\
\rowcolor{gray_table}
\textbf{MLP-finetuned}  & 10.00 & 23.10±8.17
\end{tabular}
\caption{Results of fine-tuned models with experimental setup. The LSTM-finetuned outperforms the other models in terms of the target error}
\label{tab:finetune_controller}
\end{table}

\section{Limitations and Future Work}
\label{sec:limitations}

One of the main challenges of gravitational pivoting is that to avoid extra motion of the end-effector, the object can only rotate in one direction. Hence, it is hard to recover from errors when the object moves past the target angle. To minimize the effect of the issue, we aim to rotate the object slowly, by slowly loosening the grip until some velocity is observed. However, the object can still rotate in the magnitude of 500$\degree$/s. The fast speeds necessitate accurate stopping conditions, which makes accurate target reaching very difficult. Furthermore, there is a large delay in commanding the gripper to close and the object coming to a rest, at around 0.83s. The stopping condition uses forward prediction proportional to both the delay and velocity, therefore, any errors in velocity prediction are magnified by a large time delay. A potential solution for both of these issues would be to slow the object down as it is pivoting. This could be achieved using a learning-based controller, which could make smarter decisions on when to speed up or slow down pivoting. Another potential solution would be to increase the speed of the data-collection process and in turn allow RSE-LSTM to operate at a higher frequency. This would allow more reactive control to fast changes in position and velocity. Another limitation of our work is that the performance degrades when working with unknown objects. This could be improved with collecting more varied data on a wider range of objects. Further investigation is needed at looking at the effect of out of distribution testing objects, for examples lighter objects will likely be harder to rotate. Our work aims to show that tactile information is sufficient for this problem. While it does show promising results, further work needs be done to improve the results, to make both the estimation and the controller more robust.

\newpage
\acknowledgments{We would like to thank Lily Tran for help editing the manuscript.}

\bibliography{cite}
\newpage

\section{Supplementary Material}

\subsection{Filters}

The measured angle is filtered with a Kalman filter to smooth the results. However, it was found that the Kalman filter introduces inconsistent delay to the angular velocity, hence the angular velocity is instead directly computed from the Kalman filtered angle. This angular velocity is then smoothed using a non-causal filter, with a triangular window of size 9. This filter was found to produce the smoothest results, without adding any delay. 

Following the conventions from \citep{Chen2021_Velocity}, the kinematics model used was \(x_{t+1}=Ax_{t}+w_{t}\) and the measurement model used was \(z_t=Cx_t+v_t\). Here the process noise $w_t$ and the measurement noise $v_t$ are assumed to be 0 mean with covariance matrices $Q$ and $R$ respectively. The Kalman filter aims to estimate $x_t=[\alpha, \omega]^\intercal$ using the measurements $z_t$. The initial state $x_0$ is taken to be $[z_0, 0]^\intercal$ with covariance $\Sigma_0$. The parameters used are listed in Table \ref{tab:kal_params}.

\begin{table}[htb]
\centering
\begin{tabular}{l|c}
Parameter & Value \\\hline
$A$ & $\begin{bmatrix}1 & \frac{1}{60}\\ 0 & 1\end{bmatrix}$ \\
$C$ & $\begin{bmatrix}1 & 0\end{bmatrix}$ \\
$Q$ & $\begin{bmatrix} \num{3.25e-6} & \num{6.5e-5} \\ \num{6.5e-5} & \num{1.3e-3} \end{bmatrix}$ \\
$R$ & $\num{1e-5}$ \\
$\Sigma_0$ & $\begin{bmatrix}\num{1e-5} & 0 \\ 0 & \num{1e-5} \end{bmatrix}$ \\
\end{tabular}
\caption{Kalman filter parameters}
\label{tab:kal_params}
\end{table}

\subsection{Data Collection Methodology}

Each object starts at rest on a horizontal plane. The gripper approaches the object at a global yaw angle ($\alpha_{approach}$) chosen from a set of angles $\mathrm{A}_{approach}$. The object is then grasped by closing the gripper to a fixed width for each object. This width manually is chosen such that the objects remain stationary inside the gripper while being lifted. The object is then lifted to a fixed height above the table, where an end effector yaw angle perturbation is chosen from the set of angles $\mathrm{A}_{perturb}$. The recording of both tactile measurements and the images is started after the object has been lifted. The object is held stationary for 0.5s. The gripper width is then loosened to start the gravitational pivoting. One of the two methodologies for gripper control will then be used. Once the object has finished rotating, another fixed delay of $1$s occurs, where the object is held stationary in its steady-state position. The data recording is then finished. For `Angle Goal', each parameter combination was run twice to increase the amount of data collected.

\subsection{Parameters}

\subsubsection{Data Collection}

\textbf{Rotate to stop}: 
\begin{itemize}
    \item $\mathrm{A}_{approach} = \{-30\degree, -15\degree, 0\degree, 15\degree, 30\degree\}$
    \item $\mathrm{A}_{perturb} = \{-45\degree, 0\degree, 15\degree, 30\degree, 45\degree, 60\degree\}$
\end{itemize}

\textbf{Angle Goal}
\begin{itemize}
    \item $\mathrm{A}_{approach} = \{-15\degree, 0\degree\}$
    \item $\mathrm{A}_{perturb} = \{0\degree, 30\degree, 45\degree, 60\degree\}$
    \item $\mathrm{A}_{stop} = \{15\degree, 30\degree, 45\degree\}$.
\end{itemize}

\subsubsection{Experiments}
\begin{itemize}
    \item $\mathrm{A}_{approach} = \{-30\degree, 0\degree, 30\degree\}$
    \item $\mathrm{A}_{perturb} = \{0\degree\}$
    \item $\mathrm{A}_{stop} = \{30\degree, 45\degree, 60\degree\}$
\end{itemize}

\subsubsection{Controller}

The controller was hand-tuned to maximise the performance of the system with the \textbf{Oracle} baseline. The other controllers were directly transferred using the same parameters, listed in Table \ref{tab:control_params}.

\begin{table}[htb]
\centering
\begin{tabular}{l|c}
Parameter & Value \\\hline
$\epsilon_{\alpha}$ & $1\degree$ \\
$\omega_{min}$ & $20\degree$/s \\
$t_{wait}$ & $0.75$s \\
$d$ & $0.83$s
\end{tabular}
\caption{Controller Parameters}
\label{tab:control_params}
\end{table}

\subsection{Unseen Class Experiment}

We investigate the ability of the model to generalize between different classes of objects. To do this, we train and test on different classes of objects (boxes vs cylinders). The results of this training are displayed in Table \ref{tab:unseen_class}. When training and testing on a single class, the performance of the networks is similar to training on both classes. This suggests that the network can generalize between objects of the same class. However, the performance of both the LSTM and MLP are decreased when testing on an unseen class. This is likely due to different shapes causing a large change in the resultant tactile information.

\begin{table}[htb]
\centering
\begin{tabular}{lll|rrr|rrr}
 &  &  & \multicolumn{3}{c|}{Angular Error ($\degree$)} & \multicolumn{3}{c}{Velocity Error ($\degree$/s)} \\ \cline{4-9} 
\multicolumn{1}{l}{\textbf{Methods}} & \textbf{Train} & \textbf{Test} & \multicolumn{1}{c}{\textbf{IS}} & \multicolumn{1}{c}{\textbf{DR}} & \multicolumn{1}{c|}{\textbf{SS}} & \multicolumn{1}{c}{\textbf{IS}} & \multicolumn{1}{c}{\textbf{DR}} & \multicolumn{1}{c}{\textbf{SS}} \\ \hline

MLP & Box & Box & 1.94 & 10.53 & 20.42 & 5.24 & 36.36 & 5.78\\
LSTM & Box & Box & 0.32 & 4.02 & 8.35 & 1.54 & 26.27 & 0.95\\
\rowcolor{gray_table}
MLP & Box & Cyl & 3.98 & 15.42 & 35.06 & 5.13 & 46.61 & 8.24\\
\rowcolor{gray_table}
LSTM & Box & Cyl & 3.38 & 11.99 & 18.83 & 2.949 & 44.46 & 0.72\\
MLP & Cyl & Cyl & 1.27 & 7.69 & 14.86 & 8.86 & 41.31 & 7.93\\
LSTM & Cyl & Cyl & 0.34 & 3.29 & 5.23 & 1.86 & 34.71 & 3.12\\
\rowcolor{gray_table}
MLP & Cyl & Box & 18.65 & 27.8 & 54.55 & 3.69 & 63.2 & 8.27 \\
\rowcolor{gray_table}
LSTM & Cyl & Box & 10.17 & 24.43 & 26.59 & 4.35 & 73.17 & 3.08
\end{tabular}
\caption{Results on training on an unseen class. IS, DR, SS represent the three sections of motion, Initial State, Dynamic Range and Steady State, respectively.}
\label{tab:unseen_class}
\end{table}

\subsection{Network Details and Training Details}
All models are trained using the ADAM optimizer, using the sum of both L1 and L2 functions for both $\omega$ and $\alpha$ as the loss function. We normalized the target values of $\omega$ and $\alpha$ to be close to the range [0,1] so each component of the overall loss was given equal weighting. This was found to produce the smoothest result. While training the model, data is batched by cropping sequences to be the same length as the shortest sequence in the batch. The hyperparameters, including number of layers, of both models were found using a sweep and are reported in Table \ref{tab:hyperparams}. The best model determined by the sweep was used for all experiments.  All models are trained for 100 epochs. We report the accuracy of the model in terms of the mean absolute error (MAE). 

\begin{table}[htb]
\begin{minipage}{.5\linewidth}
\begin{tabular}{l|c}
\textbf{LSTM Parameter}          & \textbf{Value}                           \\ \hline
LSTM Input Size    & 142                                                 \\
LSM Hidden Size    & 500                                                 \\
LSTM Dropout       & 0.15                                                \\
LSTM Number Layers & 3                                                   \\
MLP Layers         & 2                                                   \\
MLP Hidden Size    & 500                                                 \\
Learning Rate      & \num{5e-4}                        \\
Weight Decay       & \num{1e-6}
\end{tabular}
\end{minipage}%
\begin{minipage}{.5\linewidth}
\begin{tabular}{l|c}
\textbf{MLP Parameter}     & \textbf{Value}    \\ \hline
Input Size    & 2130     \\
Number Layers & 4        \\
Dropout       & 0.15     \\
Activation    & Tanh     \\
Learning Rate & \num{5e-4}   \\
Weight Decay  & \num{1e-6} \\
Window Size & 15
\end{tabular}
\end{minipage}
\caption{Hyperparameters used for the experiments}
\label{tab:hyperparams}
\end{table}

\subsection{Unseen Objects}

The testing results on unseen objects are displayed in Table \ref{tab:unseen_obj}. The LSTM outperformed the MLP for all metrics, however the results have degraded compared to seen objects. In particular, the MLP failed to maintain a constant value close to 0 in the angle IS, angular velocity IS and angular velocity SS sections. The LSTM still maintains predictions close to 0 for these sections, displaying the benefit of the stability of the LSTM compared to the MLP, which is much noisier with it's predictions.

To improve the generalization to unseen objects, we believe that a larger and more diverse object set will be required. This should consist of objects of a wide range of shapes, weights and surface properties. We believe that this will allow the model to learn the effects of these properties on the forces received from the tactile sensor and so better account for them.

\begin{table}[htb]
\begin{tabular}{l|lll|lll}
 & \multicolumn{3}{c|}{Angular Error ($\degree$)} & \multicolumn{3}{c}{Angular Velocity Error ($\degree$/s)}                     \\ \cline{2-7}    
     & \multicolumn{1}{c}{IS}      & \multicolumn{1}{c}{DR}      & \multicolumn{1}{c|}{SS}      & \multicolumn{1}{c}{IS}     & \multicolumn{1}{c}{DR}      & \multicolumn{1}{c}{SS}      \\ \hline

\textbf{LSTM} & \textbf{0.37±0.17} & \textbf{7.98±4.04} & \textbf{10.87±4.58} & \textbf{0.84±0.38} & \textbf{42.27±14.29} & \textbf{2.43±1.37} \\
\textbf{MLP}  & 3.49±5.14 & 13.1±4.73 & 24.49±5.92 & 5.05±3.45 & 44.44±13.52 & 5.03±1.49
\end{tabular}
\caption{Testing results on unseen objects. IS, DR, SS represent the three sections of motion, Initial State, Dynamic Range and Steady State, respectively. }
\label{tab:unseen_obj}
\end{table}

\subsection{Finetuning on real robot}

The finetuning results on the training dataset are displayed in Table \ref{tab:finetune_train}. Note that the finetune training is done following the unseen objects experiments. Although finetuning has increased error in some sections, the error within the critical `Dynamic Range' section has decreased. This section is critical in detecting when a target angle has been reached and so this decreased error reflects the decreased target error of the finetuned models.

\begin{table}[htb]
\begin{tabular}{l|lll|lll}
 & \multicolumn{3}{c|}{Angular Error ($\degree$)} & \multicolumn{3}{c}{Angular Velocity Error ($\degree$/s)}                     \\ \cline{2-7}    

     & \multicolumn{1}{c}{IS}      & \multicolumn{1}{c}{DR}      & \multicolumn{1}{c|}{SS}      & \multicolumn{1}{c}{IS}     & \multicolumn{1}{c}{DR}      & \multicolumn{1}{c}{SS}      \\ \hline
\textbf{LSTM} & \textbf{0.54±0.18} & \textbf{4.84±2.11} & \textbf{13.75±6.00} & \textbf{3.01±3.06} & 33.54±20.12 & \textbf{2.99±2.45} \\
\textbf{MLP}  & 3.08±3.89 & 7.78±3.51 & 26.32±10.25 & 4.83±4.91 & \textbf{31.36±22.30} & 5.41±3.46 
\end{tabular}
\caption{Results of training finetuned models. IS, DR, SS represent the three sections of motion, Initial State, Dynamic Range and Steady State, respectively.}
\label{tab:finetune_train}
\end{table}

\subsection{Finetuned Real-World Tracking Results}

Table \ref{tab:finetune_controller} shows the results for real-world tracking. Again these results were measured on unseen object models. While the error of real world tracking is similar to the training results, the variance is higher. This contributes to the tracking error being much higher than the average errors, with models failing at critical instances such as when the gripper width widens and the object begins to rotate faster. Such spikes in error will be smoothed out by the longer periods of slow rotations but are still present in the measurements of variance.

\begin{table}[h!]
\begin{tabular}{l|lll|lll}
 & \multicolumn{3}{c|}{Angular Error ($\degree$)} & \multicolumn{3}{c}{Angular Velocity Error ($\degree$/s)}                     \\ \cline{2-7}    

     & \multicolumn{1}{c}{IS}      & \multicolumn{1}{c}{DR}      & \multicolumn{1}{c|}{SS}      & \multicolumn{1}{c}{IS}     & \multicolumn{1}{c}{DR}      & \multicolumn{1}{c}{SS}      \\ \hline
\textbf{LSTM} & 0.69±0.61 & 4.61±4.15 & 14.74±11.82 & 3.46±3.49 & 14.51±16.59 & 5.09±14.72 \\
\textbf{MLP}  & 1.76±3.07 & 10.15±8.92 & 31.23±27.21 & 4.51±2.71 & 11.47±6.17 & 5.55±7.43
\end{tabular}
\caption{Results of finetuned models with experimental setup. IS, DR, SS represent the three sections of motion, Initial State, Dynamic Range and Steady State, respectively.}
\label{tab:finetune_controller}
\end{table}

\subsection{Test Tracking Results}

\begin{figure}[htb]
    \centering
    \includegraphics[width=1\linewidth, trim=0cm 14.4cm 40.48cm 0cm, clip]{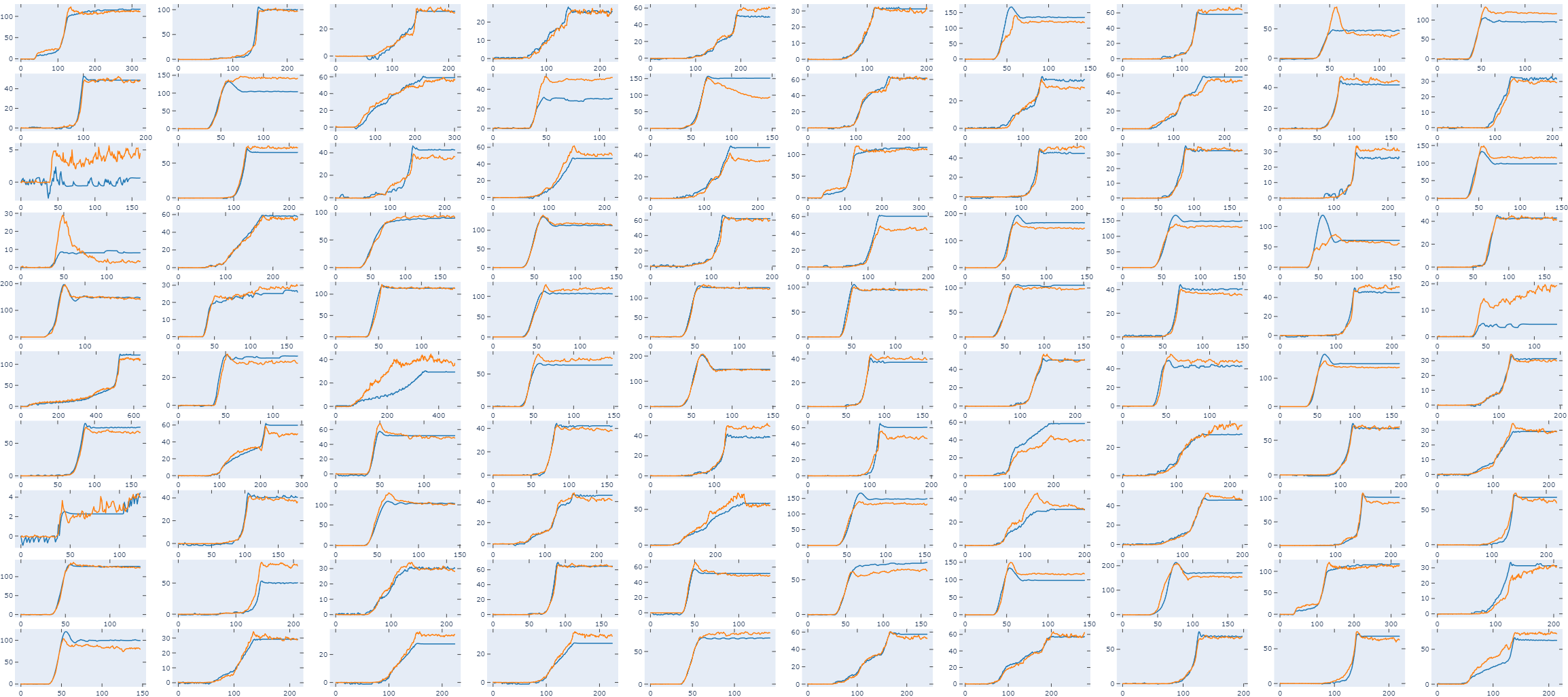}
    \caption{The LSTM tracking results on unseen data. The blue lines represent the ground truth sequences, and the orange lines represent the LSTM prediction. The y-axis represents the angular position. The scale of this axis varies between graphs.}
    \label{fig:lstm_examples}
\end{figure}

Figure \ref{fig:lstm_examples} shows more examples of the LSTM performance on unseen data (as described in \ref{subsec:rotest}). The LSTM can accurately track the object, especially when the object follows a smooth trajectory. However, the LSTM also commonly suffers from steady state errors. 

\subsection{MLP Window Size}
\label{subsec:windowexperiments}

\begin{table}[htb]
\centering

\begin{tabular}{c|ccc|ccc}
Window & \multicolumn{3}{c|}{Angular Error ($\degree$)} & \multicolumn{3}{c}{Angular Velocity Error ($\degree$/s)}                     \\ \cline{2-7}    
     \multicolumn{1}{c|}{Size} & \multicolumn{1}{c}{IS}      & \multicolumn{1}{c}{DR}      & \multicolumn{1}{c|}{SS}      & \multicolumn{1}{c}{IS}     & \multicolumn{1}{c}{DR}      & \multicolumn{1}{c}{SS}      \\ \hline

5 & 1.03 & 11.34 & 19.89 & 1.8 & 49.59 & 4.64 \\ 
15 & 1.15 & \textbf{10.47} & 18.84 & \textbf{1.75} & \textbf{40.45} & 4.23 \\
30 & 0.94 & 10.65 & 20.77 & 1.82 & 43.78 & \textbf{3.52} \\
60 & 1.34 & 11.65 & 20.66 & 1.83 & 48.04 & 6.72 \\
90 & \textbf{0.7} & 12.03 & \textbf{18.59} & 12.86 & 52.13 & 7.85 \\
\end{tabular}
\caption{Ablation study on the window size used for the MLP. IS, DR, SS represent the three sections of motion, Initial State, Dynamic Range and Steady State, respectively.}
\label{tab:window_size}
\end{table}

Table \ref{tab:window_size} shows the results for MLP models trained on an 80/20 random data split, comparing different window sizes. For Angular Error, we found that for a window size of 15, the error during Dynamic Range was lowest and had a comparable error to a window size of 90 during Steady State. We also found that a window size of 15 performed well in predicting velocity error, outpeforming all other models during the Dynamic Range stage, which is the most important for our controller.  

\subsection{Other Recurrent Models with Unseen Objects}

Table \ref{tab:other_recurrent_models} shows the results when other recurrent model types are trained to generalize to unseen objects. The GRU had identical hyperparameters to the LSTM and the RNN had the same hyperparameters expect for a dropout of 0. We found that the RNN model outperforms the MLP baseline model consistently, but performed worse than both the LSTM and GRU models. 

We also found that the LSTM and GRU models exhibit almost identical performance in all cases. The slightly lower angular error the GRU displays compared to the LSTM in the DR and SS regions comes entirely from the overperformance of the GRU on one object, Magnet. For all other objects there was minimal discrepancy between the models. We believe these models could be used interchangeably for this task.

\begin{table}[htb]

\begin{tabular}{l|lll|lll}
 & \multicolumn{3}{c|}{Angular Error ($\degree$)} & \multicolumn{3}{c}{Angular Velocity Error ($\degree$/s)}                     \\ \cline{2-7}    

     & \multicolumn{1}{c}{IS}      & \multicolumn{1}{c}{DR}      & \multicolumn{1}{c|}{SS}      & \multicolumn{1}{c}{IS}     & \multicolumn{1}{c}{DR}      & \multicolumn{1}{c}{SS}      \\ \hline
\textbf{LSTM} & \textbf{0.37±0.17} & 7.98±4.04 & 10.87±4.58 & \textbf{0.84±0.38} & 42.27±14.29 & \textbf{2.43±1.37} \\
\textbf{GRU}  &0.42±0.25	&\textbf{7.38±3.21}	&\textbf{10.69±4.31}	&1.32±0.56	&\textbf{39.51±11.87}	&3.94±1.10 \\
\textbf{RNN}  & 1.80±1.87 & 10.75±4.26 & 15.47±4.30 & 2.88±1.06 & 46.24±11.69 & 5.85±4.63
\end{tabular}
\caption{Testing other recurrent models on unseen objects. IS, DR, SS represent the three sections of motion, Initial State, Dynamic Range and Steady State, respectively.}
\label{tab:other_recurrent_models}
\end{table}

\subsection{Object Details}






\begin{table}[h!]

\begin{tabular}{c|cccccc}
 & Size [mm] & Mass [g] & Class & 3D-Printed? & Full? & Deformable \\ \hline
Toothpaste & 167 $\times$ 58 $\times$ 12  & 52 & Box & & \ForestGreencheck & \ForestGreencheck \\
Earbud & 134 $\times$ 51 $\times$ 29  & 27  & Box &  & \ForestGreencheck &  \\
Breadboard & 167 $\times$ 58 $\times$ 12  & 84  & Box &  & \ForestGreencheck &  \\
Magnet & 181 $\times$ 68 $\times$ 40 & 29 & Box &  &  \\
Deodorant & 49 $\times$ 210  & 50 & Cylinder &  &  \\
Spray2 & 41 $\times$ 158  & 135 & Cylinder & & \ForestGreencheck & \ForestGreencheck \\
Shampoo & 50 $\times$ 157  & 96 & Cylinder & \ForestGreencheck &  \\
Spray1 &  37 $\times$ 142 & 46 & Cylinder & \ForestGreencheck &  \\
Pill & 55 $\times$ 116  & 26 &  Cylinder &  &  \\
Toothbrush & 217 $\times$ 29 $\times$ 21 &  33 & Box &  & \ForestGreencheck
\end{tabular}
\caption{Details of the object distributions used in our work. The dimensions of boxes are provided in the form $length \times width \times depth$, while cylindrical objects are dimensioned as $radius \times height$}
\label{tab:objects}
\end{table} 

Table \ref{tab:objects} provides details on the properties of objects used in our work. All objects are required to be long, to allow for gravitational pivoting. The 3D-printed objects are to provide a different texture compared to the relatively smooth materials of the household objects. When objects that are partially full undergo rotation, their contents will move around such that there will be a large change in its centre of gravity. However, for this work, we only consider objects full such that the centre of gravity is relatively stable during rotation. Deformable objects include objects which can be squished such as the soft cardboard making up the toothpaste box. We tried to choose a range of objects to increase the ability for our system to generalize. 

\end{document}